\documentclass[letterpaper]{article} 
\usepackage{aaai2026}  
\usepackage{times}  
\usepackage{helvet}  
\usepackage{courier}  
\usepackage[hyphens]{url}  
\usepackage{graphicx} 
\usepackage{natbib}  
\usepackage{caption} 
\usepackage{algorithm}
\usepackage{algorithmic}

\usepackage{amsmath}
\usepackage{mathtools}
\usepackage{amssymb}
\usepackage{mdframed}
\usepackage[most]{tcolorbox}
\usepackage{multirow}
\usepackage{booktabs}

\newtcbox{\mytagred}[1][]{nobeforeafter, colframe=black, colback=white,
  boxrule=0pt, arc=2pt, boxsep=0pt, left=2pt, right=2pt, top=2pt, bottom=2pt, fontupper=\scriptsize\sf, #1}

\newtcbox{\mytaggreen}[1][]{nobeforeafter, colframe=black, colback=white,
boxrule=0pt, arc=2pt, boxsep=0pt, left=2pt, right=2pt, top=2pt, bottom=2pt, fontupper=\scriptsize\sf, #1}

\newtcbox{\mytagblue}[1][]{nobeforeafter, colframe=black, colback=white,
  boxrule=0pt, arc=2pt, boxsep=0pt, left=2pt, right=2pt, top=2pt, bottom=2pt, fontupper=\scriptsize\sf, #1}

\usepackage{newfloat}
\usepackage{listings}
\DeclareCaptionStyle{ruled}{labelfont=normalfont,labelsep=colon,strut=off} 
\floatstyle{ruled}
\newfloat{listing}{tb}{lst}{}
\floatname{listing}{Listing}
\title{Not All Tokens Are Meant to Be Forgotten}
\author{
    Xiangyu Zhou\textsuperscript{\rm 1},
    Yao Qiang\textsuperscript{\rm 2},
    Saleh Zare Zade\textsuperscript{\rm 1},
    Douglas Zytko\textsuperscript{\rm 3},
    Prashant Khanduri\textsuperscript{\rm 1},\newline
    Dongxiao Zhu\textsuperscript{\rm 1}
}
\affiliations{
    \textsuperscript{\rm 1}Department of Computer Science, Wayne State University\\
    \textsuperscript{\rm 2}Department of Computer Science and Engineering, Oakland University\\
    \textsuperscript{\rm 3}College of Innovation and Technology, University of Michigan-Flint\\

    xiangyu@wayne.edu, qiang@oakland.edu, salehz@wayne.edu, dzytko@umich.edu, khanduri.prashant@wayne.edu, dzhu@wayne.edu
}

\usepackage{bibentry}

\begin{document}

\maketitle

\begin{abstract}
Large Language Models (LLMs), pre-trained on massive text corpora, exhibit remarkable human-level language understanding, reasoning, and decision-making abilities. However, they tend to memorize unwanted information, such as private or copyrighted content, raising significant privacy and legal concerns. Unlearning has emerged as a promising solution, but existing methods face a significant challenge of over-forgetting. This issue arises because they indiscriminately suppress the generation of all the tokens in forget samples, leading to a substantial loss of model utility. To overcome this challenge, we introduce the {\textbf{Targeted Information Forgetting}} (TIF) framework, which consists of (1) a flexible targeted information identifier designed to differentiate between unwanted words (UW) and general words (GW) in the forget samples, and (2) a novel {\textbf{Targeted Preference Optimization}} approach that leverages {\em Logit Preference Loss} to unlearn unwanted information associated with UW and {\em Preservation Loss} to retain general information in GW, effectively improving the unlearning process while mitigating utility degradation. Extensive experiments on the TOFU and MUSE benchmarks demonstrate that the proposed TIF framework enhances unlearning effectiveness while preserving model utility and achieving state-of-the-art results.
\end{abstract}

\begin{links}
    \link{Code}{https://github.com/xzhou98/Unlearning-TPO}
\end{links}

\section{Introduction}
Large Language Models (LLMs), pre-trained on vast text corpora, demonstrate strong capabilities in text generation and nuanced language understanding~\cite{brown2020language}. However, they often memorize parts of their training data~\cite{carlini2021extracting}, which, while useful for tasks such as question answering~\cite{brown2020language} and code generation~\cite{jiang2024survey}, raises security and safety concerns. Specifically, memorization of personally identifiable information (PII) or copyrighted content poses risks of privacy violations or copyright infringement~\cite{carlini2021extracting,karamolegkou2023copyright,li2024wmdp,zhou2023hijacking,zhou2024learning,zade2025automatic}. To address these risks, recent work has explored machine unlearning techniques for LLM dememorization.

\begin{figure*}[t]
    \centering
    \includegraphics[width=0.93\linewidth]{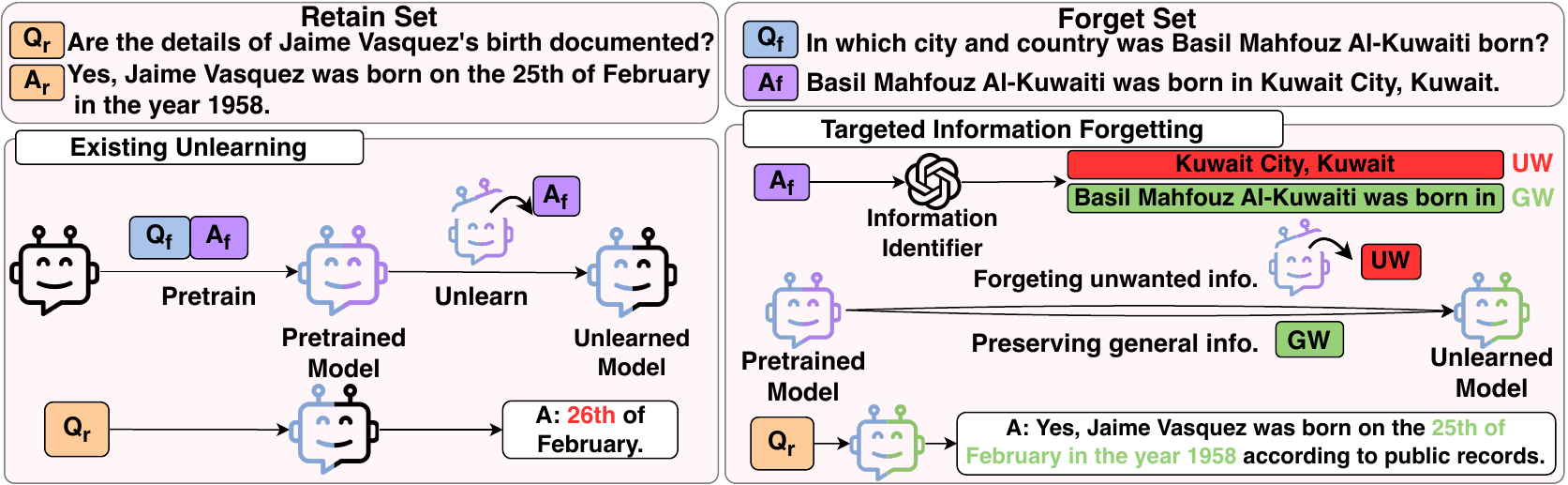}
    \caption{ {\bf Illustration of the proposed TIF framework}. TIF exploits an unwanted information identifier to differentiate between unwanted and general information in the forget sample (e.g., $\text{A}_\text{f}$ in the right panel). The former is represented by Unwanted Words (UW) and the latter by General Words (GW). Instead of removing the entire response $\text{A}_\text{f}$, TIF selectively unlearns only UW while preserving general knowledge associated with GW by retraining on GW. This targeted approach enables effective forgetting while maintaining model utility. The right panel demonstrates a more complete and correct model response compared to the existing unlearning approach on the left.
    }
    \label{fig:illustration}
\end{figure*}

Machine unlearning~\cite{cao2015towards,bourtoule2021machine,nguyen2022survey} was developed as an efficient approach to remove the influence of specific training samples from pre-trained models, eliminating the need for full retraining. Recently, unlearning techniques have been adapted for LLMs, raising ongoing challenges in precisely removing private or copyrighted content learned from specific training samples ~\cite{yao2023large,pawelczyk2023context,eldan2023s,chen2023unlearn}. Early approaches mainly rely on fine-tuning pre-trained models~\cite{li2024wmdp,chen2023unlearn,si2023knowledge,jang2022knowledge}, with some using gradient ascent (GA) optimization to achieve the goal of unlearning~\cite{yao2023large}. However, due to the unbounded nature of the loss function, GA-based methods lack precise control over updates, often leading to catastrophic collapse, where the model’s overall performance deteriorates~\cite{zhang2024negative}. To address this issue, recent work introduces preference optimization-based frameworks such as Negative Preference Optimization (NPO) \cite{zhang2024negative} and SimNPO \cite{fan2024simplicity} to mitigate performance collapse.

\begin{figure}[t]
    \centering
    \includegraphics[width=1\linewidth]{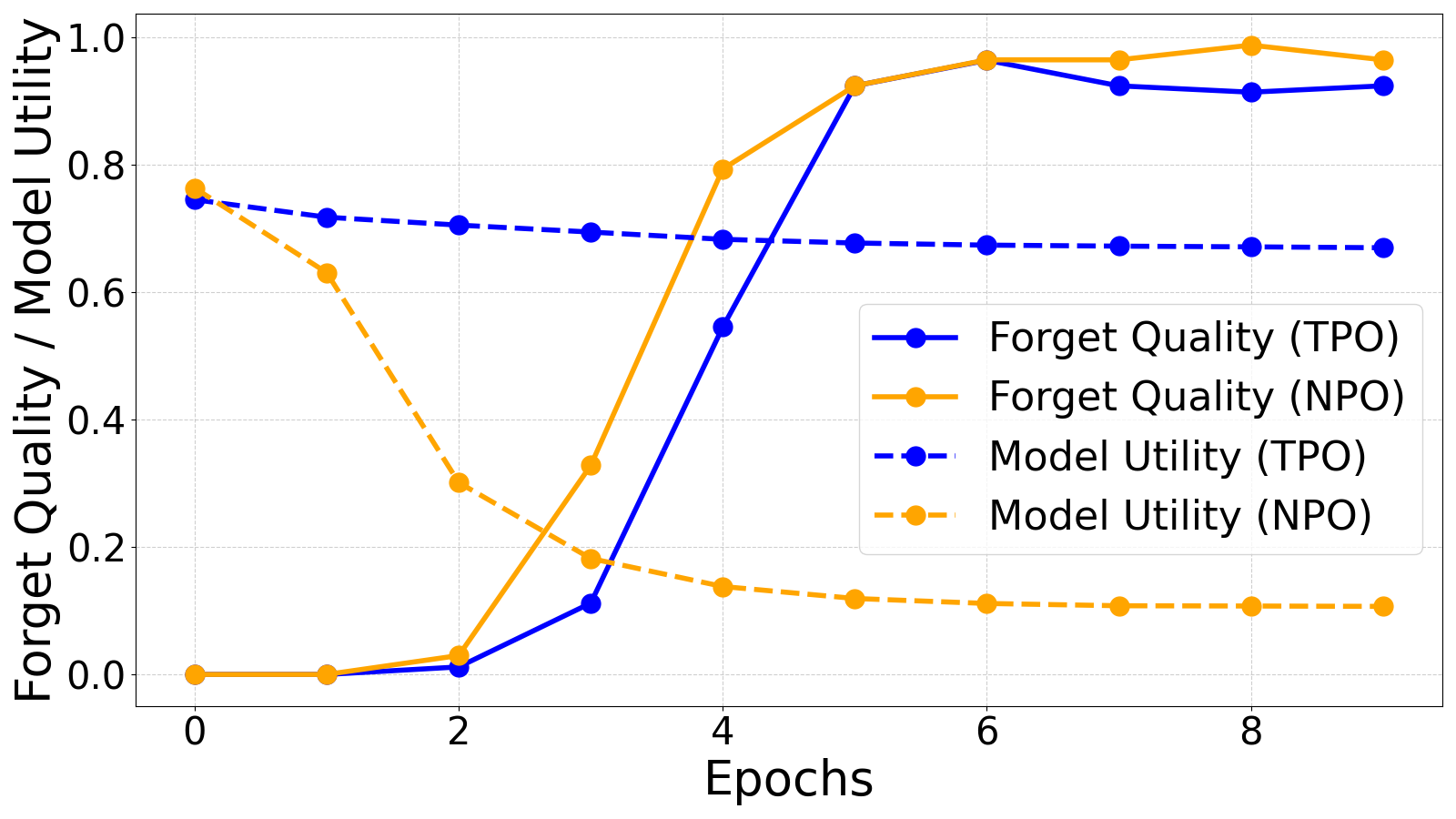}
      \caption{Comparison of our TPO and NPO \cite{zhang2024negative} on key metrics: forget quality and model utility. The results are derived from the Forget05 task of the TOFU dataset \cite{maini2024tofu}.}
    \label{fig:Comparison}
\end{figure}

Despite these foundational efforts, LLM unlearning still faces several critical challenges: 
\textbf{(C1) Ambiguous Unlearning Targets}\label{label:C1}. Most existing approaches treat the entire forget sample as the unlearning target without differentiating between unwanted information (to be unlearned) and general information (to be retained), as shown in Figure~\ref{fig:illustration}. This lack of distinction often leads to significant degradation of model utility \cite{wang2024selective,liu2024large,lynch2024eight,rezaei2024restor}. 
\textbf{(C2) Lack of Flexible and Generalizable Unwanted Information Identification}. Recent methods attempt fine-grained unlearning but face critical limitations in information identification: ECO~\cite{liu2024large} employs a sentence-level identifier that overfits to specific keywords in the forget sample (e.g., the ``college''), rather than aligning with the unlearning requester's intent \cite{thaker2024position}. This leads to insufficient unlearning when prompts containing these specific keywords are removed. While SEUL~\cite{wang2024selective} improves precision by leveraging generative models (e.g., ChatGPT) to identify continuous sensitive spans (e.g., PII), it remains limited to handling diverse unlearning targets (e.g., copyrighted content). This rigidity in identification compromises both effectiveness and generalizability. \textbf{(C3) Sensitivity to Forget Set Size}. Methods based on preference optimization~\cite{zhang2024negative,rafailov2024direct} mitigate catastrophic collapse more effectively than other baselines, helping to preserve model utility. However, their effectiveness declines significantly as the forget set size increases, resulting in notable utility loss~\cite{liu2024large}, as shown in Figure~\ref{fig:Comparison}. 

\begin{figure}[h]
    \centering
    \includegraphics[width=1\linewidth]{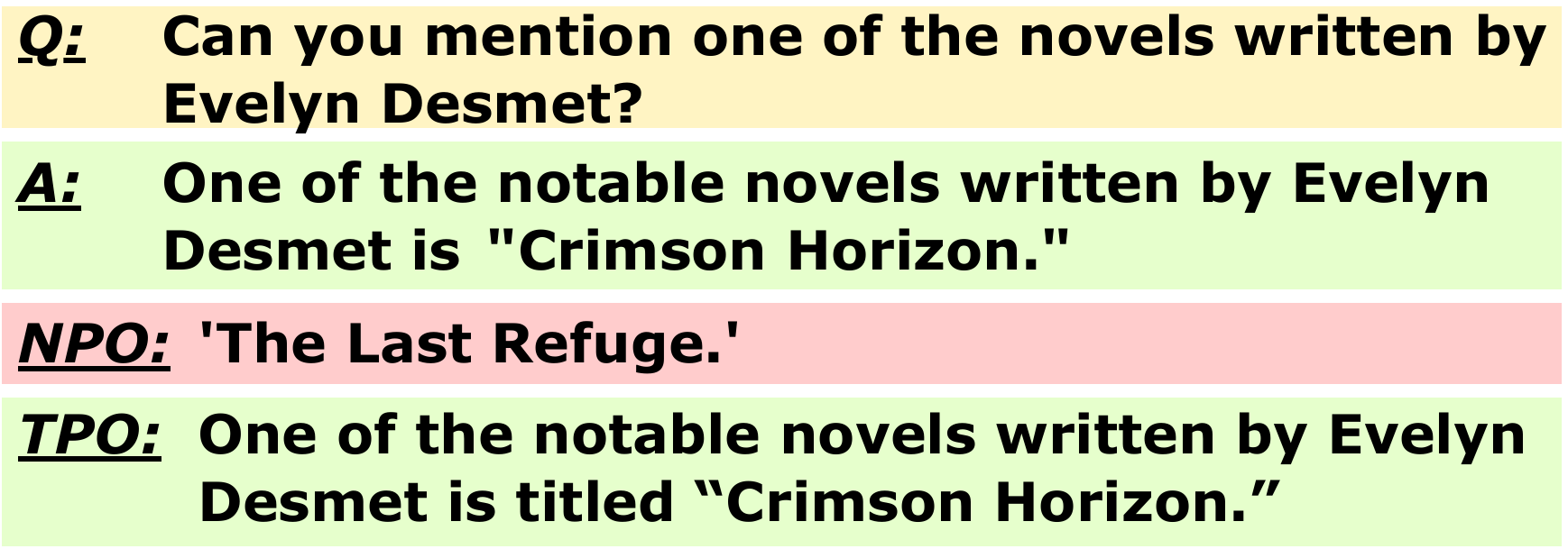}
      \caption{Illustration of responses generated by models unlearned using TPO (ours) and NPO on the retain set. The models were unlearned on the Forget05 task of the TOFU.}
    \label{fig:readability}
\end{figure}

To address the challenges {\bf (C1)-(C3)}, we propose the \textbf{Targeted Information Forgetting (TIF)} framework, as illustrated in Figure~\ref{fig:illustration}. Our main contributions are listed below:

\textbf{(1) TIF Framework.} To tackle {\bf (C1)}, we propose a novel TIF framework for LLM unlearning. Different from existing unlearning approaches such as NPO, which predominantly unlearn entire information associated with the forget instances (e.g., $\bf{A_f}$ in the left panel of Figure~\ref{fig:illustration}), our TIF is designed to unlearn only the targeted unwanted information, such as the city of born in the right panel’s example. General information is often associated with some ``General Words ({\bf GW})'', including stop words and commonly used phrases, which frequently appear in both retain and forget sets. In contrast, ``Unwanted Words ({\bf UW})'' correspond to specific private or copyrighted content, such as city of born. By specifically targeting only UW for unlearning, our TIF preserves more general information compared to existing methods like NPO, effectively preventing over-forgetting and enabling the model to generate more readable responses, as demonstrated in the retain set answers in Figure~\ref{fig:illustration}.

\textbf{(2) Unwanted Information Identification.} To address {\bf (C2)}, we develop flexible yet effective approaches for unwanted information identification: a generative model such as ChatGPT-4, and a discriminative model such as DistilBERT~\cite{sanh2019distilbert}, to effectively differentiate UW from GW. We evaluate their unlearning performance and illustrate their respective use cases. As a bottom line, even identifying function words (e.g., the, is, or an) as GW according to linguistics would improve model utility preservation.

\begin{figure*}[t]
    \centering
    \includegraphics[width=0.9\linewidth]{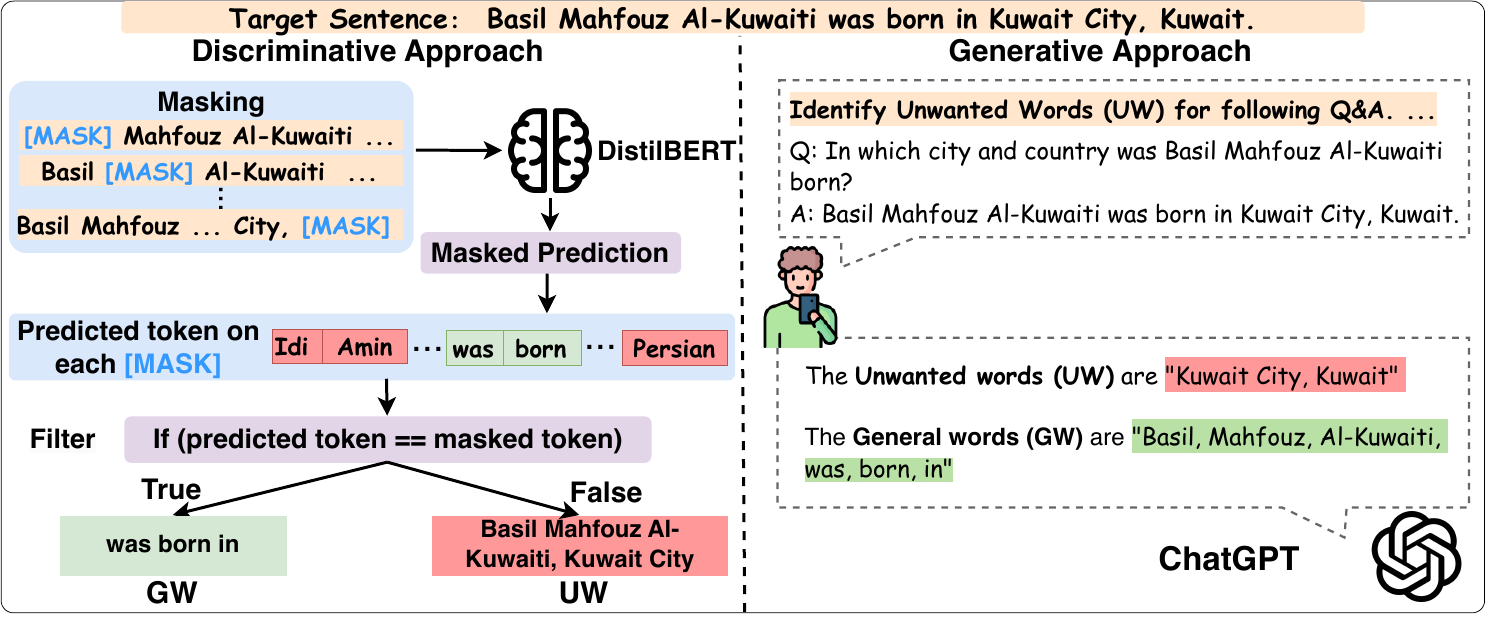} 
    \caption{ Illustration of the proposed information identification. The discriminative approach (left) uses DistilBERT with masked token predictions, while the generative approach (right) leverages ChatGPT with explicit instructions to identify UW and GW.} 
    \label{fig:pipline}
\end{figure*}

\textbf{(3) A Novel Optimization Method to Retain Model Utility.} To overcome {\bf (C3)}, we advance preference optimization algorithms ~\cite{zhang2024negative,fan2024simplicity} by introducing Targeted Preference Optimization (\textbf{TPO}), a novel optimization objective designed to mitigate the significant utility degradation observed in NPO. Specifically, our TPO integrates two innovative components: {\em Preservation loss (PL)} to maintain general model utility by retraining on GW, and {\em Logit preference loss (LPL)} to unlearn unwanted information in UW. This optimization approach effectively balances general information retention and unwanted information forgetting, improving the robustness of preference-guided optimization even with larger forget sets. As shown in Figure \ref{fig:Comparison}, our approach, TPO, achieves a comparable forget quality to NPO while significantly preserving a higher model utility. This allows the model to generate accurate information for answers in the retain set. In contrast, NPO struggles to retain essential knowledge from the retain set, as illustrated in Figure~\ref{fig:readability}.

\section{Problem Formulation}
\subsection{LLM Unlearning}
\label{subsec: LLMunlearning}
LLM unlearning aims to remove the influence of data points $\xi_f \coloneqq (x_f, y_f) \sim \mathcal{D}_f$, while preserving the integrity of the remaining knowledge in the model. Given an original model $\mathcal{M}_{\boldsymbol{\theta}_0}$ trained on a dataset $\mathcal{D}$, the goal is to unlearn $\mathcal{D}_f \subset \mathcal{D}$, which represents the subset of data points that must be forgotten. Furthermore, we define $\xi_r \coloneqq (x_r, y_r) \sim \mathcal{D}_r$, where $\mathcal{D}_r = \mathcal{D} \setminus \mathcal{D}_f$ as the retain set, which consists of data points whose information must be preserved. The objective is to update the model parameters $\boldsymbol{\theta}$ such that the unlearned model $\mathcal{M}_{\boldsymbol{\theta}}$ no longer generates $y_f$ in response to $x_f$ while maintaining its original performance on $\mathcal{D}_r$.

To achieve this goal, the unlearning procedure incorporates a forgetting objective on $\mathcal{D}_f$ and a retention objective on $\mathcal{D}_r$. Formally, the unlearning process is defined as~\cite{yao2023large,fan2024simplicity}:
    \begin{align}
  \min_{\boldsymbol{\theta}} \mathbb{E}_{\xi_f \sim \mathcal{D}_f}[\ell_{f}(y_f|x_f; \boldsymbol{\theta})] + \mathbb{E}_{\xi_r \sim \mathcal{D}_r}[\ell_r(y_r|x_r;\boldsymbol{\theta})],
\end{align}
where $\ell_f$ and $\ell_r$ represent the forget and retain losses, respectively. Specifically, the forget loss $\ell_f$ determines how well the model $\mathcal{M}_{\boldsymbol{\theta}}$ suppresses the association between $x_f$ and $y_f$, ensuring unwanted information is unlearned. Meanwhile, the retain loss $\ell_r$ enhances the model’s ability to maintain accurate associations between $x_r$ and $y_r$, preserving its original performance on $\mathcal{D}_r$.

\subsection{Targeted Unlearning}
As discussed earlier, a majority of works~\cite{si2023knowledge,chen2023unlearn,zhang2024negative,fan2024simplicity} have consistently treated the entire token sequence $y_f$ as the unlearning target for each sample $\xi_f \coloneqq (x_f,y_f)$  in the forget set $\mathcal{D}_f$, overlooking a critical question central to the process of LLM unlearning.
\begin{mdframed}[]
\begin{center}
\textbf{\emph{Are all the words in the forget sample essential for unlearning in LLMs?}}
\end{center}
\end{mdframed}

We hypothesize that ``{\bf \em Only certain words in the forget samples are relevant to the unlearning target, while others are crucial for maintaining the model's general utility.}'' To test this hypothesis, we refine the unlearning objective to focus on forgetting only certain UW, rather than the entire sequence $y_f$. We decompose $y_f$ into $\hat{y}$ and $ \bar{y}$, where $\hat{y}$ represents UW containing unwanted (e.g., private or copyrighted) information that must be forgotten, and $\bar{y}$ represents GW carrying general information (e.g., common or stop words). \\
Notably, some tokens in $\bar{y}$ may overlap with those in $y_r$, introducing general information in the samples in $\mathcal{D}_f$ shared with $\mathcal{D}_r$. Unlearning the entire $y_f$ may also unintentionally remove shared information in $\bar{y}$, leading to a decline in the model's performance on the retain set. Therefore, we emphasize that unlearning should exclusively target UW $\hat{y}$, ensuring that only the necessary information is unlearned while preserving general information. The refined targeted unlearning objective is formulated as:
    \begin{align}
    \label{eq:2}
   \min_{\boldsymbol{\theta}} \mathbb{E}_{\xi_f \sim \mathcal{D}_f}[\ell_{f}(\hat{y}|x_f; \boldsymbol{\theta})] + \mathbb{E}_{\xi_r \sim \mathcal{D}_r}[\ell_r(y_r|x_r;\boldsymbol{\theta})],
\end{align}
where $ ~y_f = \hat{y} \cup \bar{y}$.

\section{Targeted Information Forgetting (TIF)}
To achieve effective unlearning while maintaining model utility, we introduce a two-stage framework: (1) An {\em information identifier} to differentiate between UW and GW in the unlearning samples. (2) A {\em novel objective, TPO,} that refines UW logits while retraining on GW, ensuring efficient unlearning without compromising utility.

\subsection{Unwanted Information Identification\label{sec:identification}}
We investigate unwanted information identification through two distinct approaches, utilizing {\em discriminative} and {\em generative} language models (LMs).

\subsubsection{Discriminative Encoder-Only LM.\label{sec:ds_approach}} To detect unwanted information for unlearning tasks, we utilize an encoder-only LM, DistilBERT~\cite{sanh2019distilbert}, denoted as $\mathcal{M}_{\text{bert}}$. This method leverages the contextual encoding of masked LMs to estimate the likelihood of each masked word, allowing differentiation between GW and UW. Firstly, given a sample $\xi_f \coloneqq (x_f, y_f)$ from the forget set, where $y_f = [w_1, \cdots, w_i, \cdots, w_n]$ is a word sequence, we sequentially replace each word $w_i$ in $y_f$ with a special \texttt{[MASK]} token. This transformation produces a masked sequence $y'_{f_i} = [w_1, \dots, w'_i, \dots, w_n]$, where $w'_i = \texttt{[MASK]}$, as illustrated in the left panel of Figure~\ref{fig:pipline}. Next, the masked sequence $y'_{f_i}$ is fed into $\mathcal{M}_{\text{bert}}$ along with $x_f$ to predict the masked token, formally: $w^{\text{pred}}_i = \mathcal{M}_{\text{bert}}(x_f, y'_{f_i})$. If the predicted masked token matches the original masked word, $w_i$ is labeled as GW, indicating general information. Otherwise, $w_i$ is marked as UW for target unlearning.

\subsubsection {Generative Decoder-Only LM.}
To harness the power of generative decoder-only LMs in capturing contextual and semantic information from text, we employ ChatGPT-4 to directly distinguish between UW and GW by analyzing the semantics of $y_f$, as shown in the right panel of Figure~\ref{fig:pipline}. Detailed task instructions can be found in Table \ref{tab:instruction_gpt} in the Appendix \ref{sec:instruction_gpt}. Furthermore, we also present a detailed comparison of discriminative and generative approaches in relation to unlearning performance in Appendix~\ref{subsec:ga_vs_da}.

\begin{figure}[h]
    \centering
    \includegraphics[width=1\linewidth]{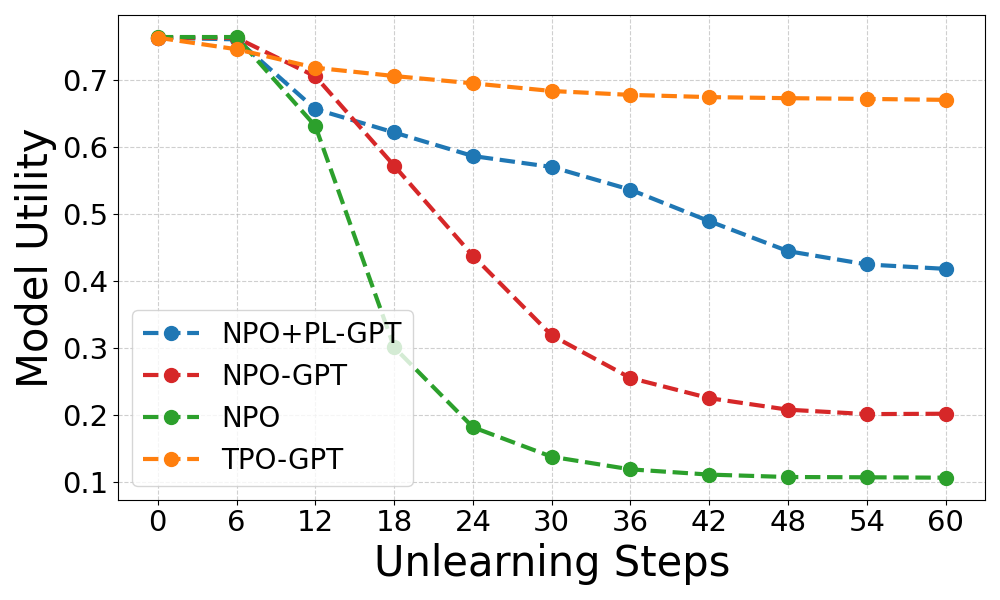}
    \caption{ Model utility across various methods on the TOFU forget05 task. Each line represents evaluations conducted at every epoch (6 steps). ``\textendash GPT'' denotes the use of ChatGPT-4 for unwanted information identification, while ``PL'' refers to the approach plus the PL Loss.}
    \label{fig:motivation2_a}
\end{figure}

\begin{figure}[h]
    \centering
    \includegraphics[width=1\linewidth]{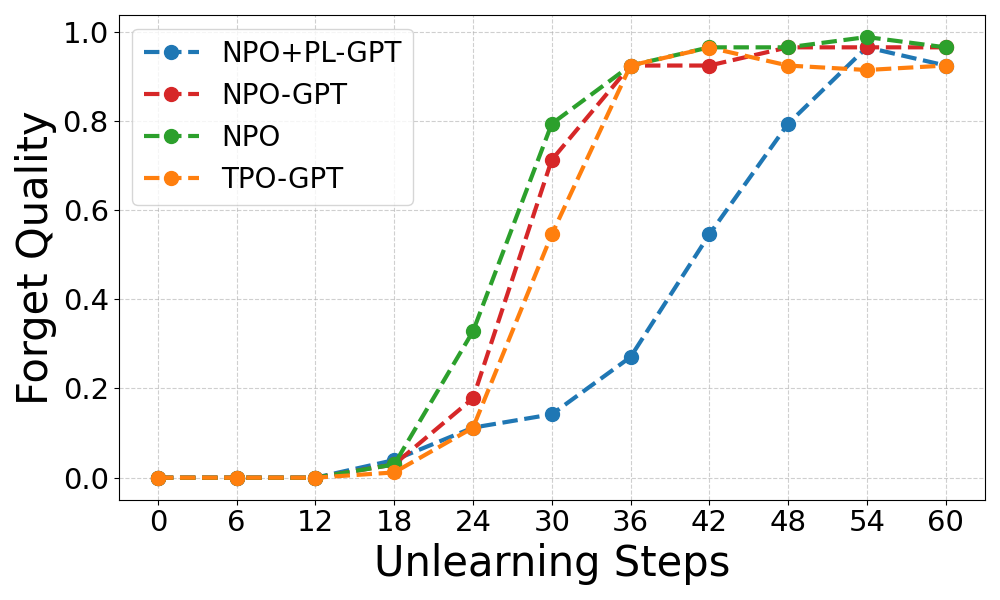}
    \caption{ Forget quality across various methods on the TOFU forget05 task. Each line represents evaluations conducted at every epoch (6 steps). ``\textendash GPT'' denotes the use of ChatGPT-4 for unwanted information identification, while ``PL'' refers to the approach plus the PL Loss.}
    \label{fig:motivation2_b}
\end{figure}

\subsection{Targeted Preference Optimization (TPO)}\label{sec:TPO}

\textbf{Motivation.} Although numerous unlearning methods, such as NPO~\cite{zhang2024negative}, have demonstrated strong performance on benchmarks such as TOFU~\cite{maini2024tofu} and MUSE~\cite{shi2024muse}, most of them struggle with model utility degradation when handling large forget sets~\cite{liu2024large}. Specifically, we evaluate the model utility and forgetting quality of NPO on TOFU, as illustrated in Figures~\ref{fig:motivation2_a} and~\ref{fig:motivation2_b}. A significant decline in model utility is evident, with the score dropping from 0.76 to 0.11, highlighting a severe utility degradation issue. We hypothesize that this degradation stems from NPO's indiscriminate handling of the entire forget samples as unlearning targets, failing to differentiate between unwanted and general information. 

To validate this, we integrate the unwanted information identifier into NPO, referred to as NPO-GPT in Figure~\ref{fig:motivation2_a}. While NPO-GPT achieves a higher model utility score compared to the standard NPO, it still experiences a 74\% decline in utility. These results suggest that merely incorporating an information identifier into NPO is insufficient to mitigate utility degradation significantly. To address this limitation, we propose TPO, a novel optimization approach designed to maintain model utility while ensuring effective unlearning.

\begin{figure}[t]
    \centering
    \includegraphics[width=\linewidth]{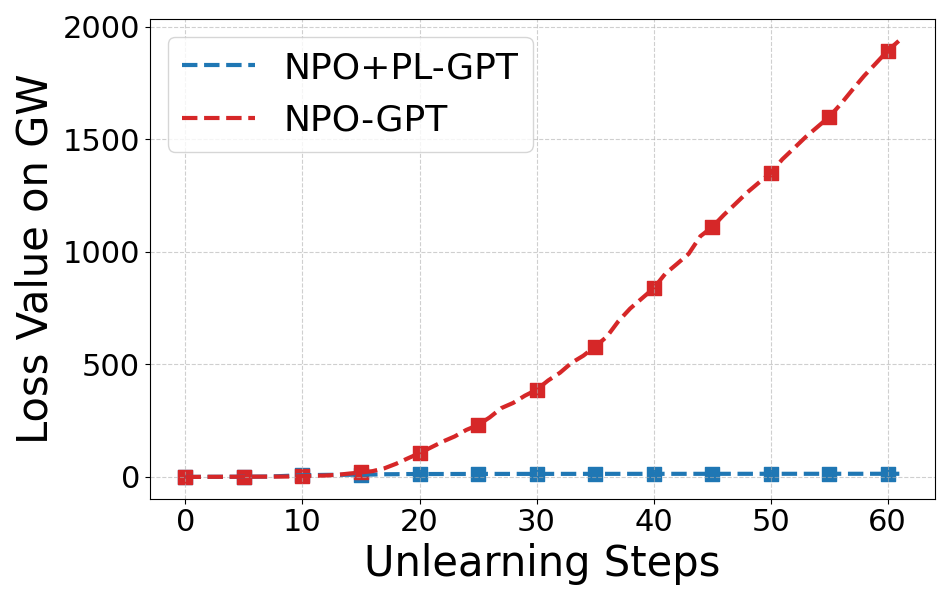}
    \caption{ Evaluation of cross-entropy loss values on GW for NPO-GPT and NPO-GPT+PL at each unlearning step.  All results are obtained for the Forget05 task in the TOFU dataset, with models trained over 10 epochs.}
    \label{fig:motivation1_a}
\end{figure}

\textbf{Preservation Loss (PL).}
To further investigate the reason for the model utility degradation observed in NPO-GPT, we analyze the cross-entropy loss values for GW, while the unlearning is limited to UW, as defined in Equation~\eqref{eq:2}. Although the model is not explicitly optimized to forget GW, the increase in loss values for NPO-GPT indicates that GW are also forgotten, as shown in Figure \ref{fig:motivation1_a}. This observation naturally leads to our key idea: how can we prevent the forgetting effects on GW and, in turn, keep their loss values as low as possible during unlearning optimization? To address this, we introduce the Preservation Loss (PL), which integrates a cross-entropy loss term on GW to explicitly prevent the model from forgetting general information, formally:
    \begin{equation}
 \textstyle   \ell_{\text{PL}}(\boldsymbol{\theta}) = -\mathbb{E}_{\xi_f\sim D_f}\big[ \log P_{\boldsymbol{\theta}} (\bar{y}|x_f)\big],
\end{equation}
where $\bar{y}$ represents the GW.

To validate the effectiveness of the PL term, we integrate it into NPO-GPT, forming NPO+PL-GPT, and evaluate its performance. As illustrated in Figure \ref*{fig:motivation1_a}, incorporating PL helps maintain stable and low loss values for GW. Consequently, NPO+PL-GPT exhibits a significantly slower decline in model utility while achieving comparable forget quality to NPO and NPO-GPT, as shown in Figures~\ref{fig:motivation2_a} and~\ref{fig:motivation2_b}. These initial results demonstrate that PL effectively mitigates model utility degradation, particularly for preserving the general information we aim to retain.

\begin{figure}[t]
    \centering
    \includegraphics[width=\linewidth]{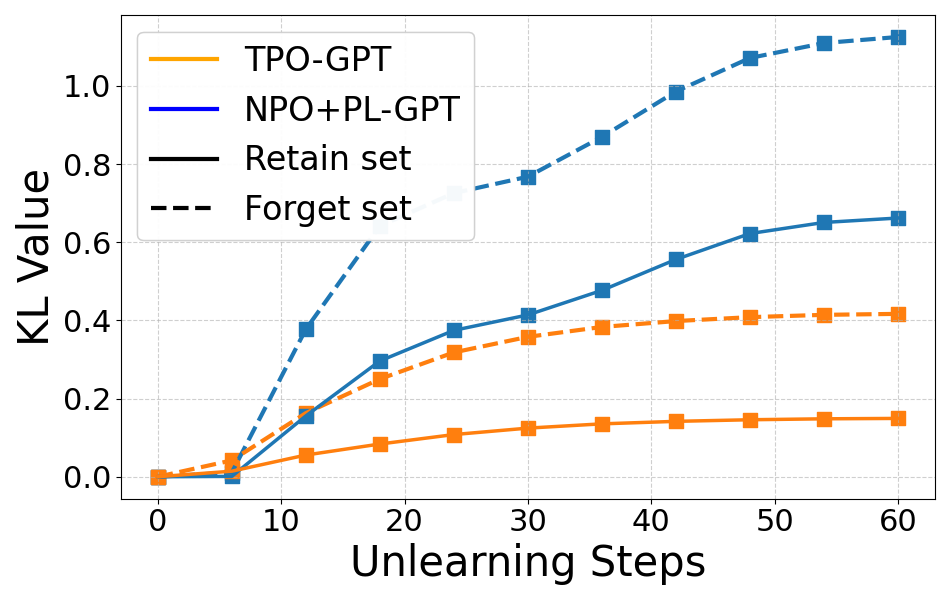}
    \caption{ The KL divergence between the reference model and the unlearned models on both forget and retain sets.  All results are obtained for the Forget05 task in the TOFU dataset, with models trained over 10 epochs.}
    \label{fig:motivation1_b}
\end{figure}

\textbf{Logit Preference Loss (LPL).} 
Upon further examination of Figure~\ref{fig:motivation2_a}, we observe that model utility degradation persists even after incorporating unwanted information identification (GPT) and PL into NPO, as seen in NPO+PL-GPT. We hypothesize that this issue stems from the limitations of NPO itself. Specifically, the unlearning process in NPO likely introduces excessive changes from the original model $\mathcal{M}_{\boldsymbol{\theta}_\text{o}}$, which serves as a reference model with parameters kept frozen during unlearning, to the final unlearned model $\mathcal{M}_{\boldsymbol{\theta}}$, particularly affecting certain general information. 

To further validate this hypothesis, we analyze the logit distribution by computing the KL divergence between $\mathcal{M}_{\boldsymbol{\theta}_\text{o}}$ and $\mathcal{M}_{\boldsymbol{\theta}}$ for both the forget and retain sets, as shown in Figure~\ref{fig:motivation1_b}. The high KL divergence observed in both sets suggests that $\mathcal{M}_{\boldsymbol{\theta}}$ (NPO-GPT+PL) has unintentionally forgotten not only the unwanted information but also general information. Specifically, NPO functions by directly reducing the probability assigned to target tokens, which is computed using the softmax function: $P(y_t) = \frac{\exp{(z_t)}}{\sum^V_{j=1}\exp(z_j)}$, where the $V$ represents the vocabulary size, and $z_t$ denotes the logit for target token $y_t$. However, reducing the probability of target token $P(y_t)$ can be achieved not only by decreasing its logits $z_t$ but also by increasing the logits of other tokens in the vocabulary. This unintended effect distorts the model's overall logit distribution, potentially compromising its ability to retain general information. The key challenge is to develop a new optimization strategy for the target tokens (UW) that selectively impacts their logit distribution while preserving the general information from GW.

To tackle this challenge, we introduce Logit Preference Loss (LPL), which takes over NPO in suppressing unwanted information during unlearning, as:
    \begin{equation}
    \ell_{\text{LPL}}(\boldsymbol{\theta}) = -\mathbb{E}_{\xi_f\sim D_f}\bigg[ \frac{2}{\beta}\log\sigma \bigg(\beta \frac{1}{|\hat{y}|} \sum_{i=1}^{|\hat{y}|} (z_i^{\boldsymbol{\theta}_\text{o}} - z_i^{\boldsymbol{\theta}}) \bigg)\bigg],
\end{equation}
where $z_i$ here denotes the logit of target token $\hat{y}_i$, and $\boldsymbol{\theta}$ and $\boldsymbol{\theta}_\text{o}$ represent the parameters of the unlearned model $\mathcal{M}_{\boldsymbol{\theta}}$ and the original model $\mathcal{M}_{\boldsymbol{\theta}_\text{o}}$, respectively. 

Different from NPO, LPL explicitly reduces only the logits of target tokens (UW) by enforcing a preference loss between $\mathcal{M}_{\boldsymbol{\theta}}$ and $\mathcal{M}_{\boldsymbol{\theta}_\text{o}}$. The primary function of LPL is to maximize the difference in target token logits between $\mathcal{M}_{\boldsymbol{\theta}}$ and $\mathcal{M}_{\boldsymbol{\theta}_\text{o}}$, while preserving the overall logit distribution. This targeted approach ensures that only the unwanted information associated with the target tokens is unlearned, without affecting general information across other tokens. As a result, LPL enables a more precise unlearning process while significantly improving model utility retention.

In summary, our proposed approach, \textbf{Targeted Preference Optimization (TPO)}, for targeted unlearning is formulated as:
    \begin{align}
    \begin{split}
\mathbb{E}_{\xi_f\sim D_f}\bigg[ &-\tfrac{2}{\beta}\log\sigma\big(\beta\,(z_{\boldsymbol{\theta}_\text{o}}(\hat{y}|x_f) - z_{\boldsymbol{\theta}}(\hat{y}|x_f))\big)  \\
& - \log P_{\boldsymbol{\theta}}(\bar{y} \mid x_f)
\bigg],
\end{split}
\end{align}
where LPL is applied to unlearn the unwanted information associated with UW ($\hat{y}$) and PL is used to preserve general information in  GW ($\bar{y})$.

Finally, as the initial results shown in Figure~\ref{fig:motivation1_b}, our approach TPO-GPT minimizes the disruption to the logit distribution on both the forget and retain sets compared to NPO+PL-GPT. Furthermore, TPO-GPT maintains most of the model utility while achieving a comparable level of forget quality to the NPO-based methods, as shown in Figures~\ref{fig:motivation2_a} and \ref{fig:motivation2_b}. 

\section{Experimental Setting \label{sec:setting}}

\subsection {Datasets and Metrics} 

We evaluate the proposed approach alongside the baseline methods on the two widely used benchmark datasets: MUSE~\cite{shi2024muse} and TOFU~\cite{maini2024tofu}. 

\textbf{(1) MUSE} is a benchmark for unlearning the copyrighted content with two unlearning tasks: forgetting the Harry Potter books (termed `Books’) and news articles (termed `News’), respectively. To evaluate the effectiveness of unlearning and the preservation of utility for MUSE, we use three metrics: Verbatim Memorization (VerbMem), Knowledge Memorization (KnowMem), and Privacy Leakage (PrivLeak). VerbMem and KnowMem are measured using ROUGE-L F1~\cite{rouge2004package}, where lower scores indicate reduced verbatim and factual memorization, respectively. PrivLeak quantifies privacy risks using the Min-K\% Prob metric~\cite{shi2023detecting} in a membership inference attack. A value close to zero indicates minimal privacy leakage, while large positive/negative values suggest over-/under-forgetting. We conduct our experiments on MUSE using ICLM-7B~\cite{shi2023context} and LLaMA-2 7B~\cite{touvron2023llama}.

\textbf{(2) TOFU} is a synthetic Q\&A dataset of 200 author biographies with three unlearning tasks: forget 1\%, 5\%, and 10\% of the author profiles. We evaluate unlearning performance using two key metrics: Forget Quality and Model Utility as defined in~\cite{maini2024tofu}. Forget quality is quantified using the $p$-value from a Kolmogorov-Smirnov (KS) test, where a higher $p$-value indicates greater similarity between the output distributions of the unlearned and the retained model. The retained model\label{lab:retained_model} denotes retraining an LLM from scratch on the retain dataset while excluding the forget set and is regarded as the gold standard for unlearning~\cite{maini2024tofu,zhang2024negative}. Model utility measures the model's performance on the retain set and its ability to retain real-world knowledge. This is assessed using various metrics, including ROUGE-L~\cite{rouge2004package} and Truth Ratio~\cite{maini2024tofu}. Experiments on TOFU utilize LLaMA-2 7B and LLaMA-3.2 3B~\cite{dubey2024llama}.

The LLMs and the evaluation metrics across unlearning benchmarks are summarized in Table~\ref{tab:summary} (Appendix).

\subsection {Unlearning Baselines}
We compare our method with baselines, i.e., GA, NPO, and SimNPO, on both MUSE and TOFU. For other baselines, such as Task Vector for MUSE and Kahneman-Tversky Optimization (KTO) for TOFU, we strictly follow their original implementations outlined in their respective benchmarks. We also evaluate the impact of incorporating Gradient Descent on the retain (GDR) loss with the baselines, i.e., GA\textsubscript{GDR}, NPO\textsubscript{GDR}, SimNPO\textsubscript{GDR}, and TPO\textsubscript{GDR}, on MUSE. Specifically, the GDR loss~\cite{liu2022continual,yao2023large,zhang2024negative,shi2024muse} is a standard gradient descent objective applied to the cross-entropy loss on the retain set $\mathcal{D}_r$. This approach enables the model to be explicitly trained to maintain performance on the retain set $\mathcal{D}_r$. More details of all baseline methods are provided in Appendix~\ref{sec:baselines}.

\subsection {Unwanted Information Identifier}
We employ two different unwanted information identifiers for TOFU dataset: a generative LM using ChatGPT-4o (via the web interface) and a discriminative LM using DistilBERT (Section~\ref{sec:ds_approach}). In Appendix~\ref{subsec:ga_vs_da}, we further examine the effectiveness of unlearning methods using both identifiers, showing that the generative LM approach enables a better balance between forget quality and model utility compared to the discriminative LM approach. For the MUSE dataset, which is non–QA in nature, we treat the beginning-of-sequence token \texttt{<bos>} as the input $x$, and define the remaining tokens as the target sequence $y$. Because each sample in the forget set of the MUSE Books dataset contains approximately 175k words (more than 200k tokens), whereas current GPT models, including ChatGPT-4o, can only handle a maximum token window size of 128k tokens (roughly 100k words). Therefore, the GPT models cannot process all the information from individual samples. It is challenging to achieve stable and consistent UW identification with GPT models for the MUSE dataset. We thus only adopt the discriminative LM approach as the unwanted information identifier on this dataset.

\begin{figure*}[t]
    \centering
    \includegraphics[width=0.97\linewidth]{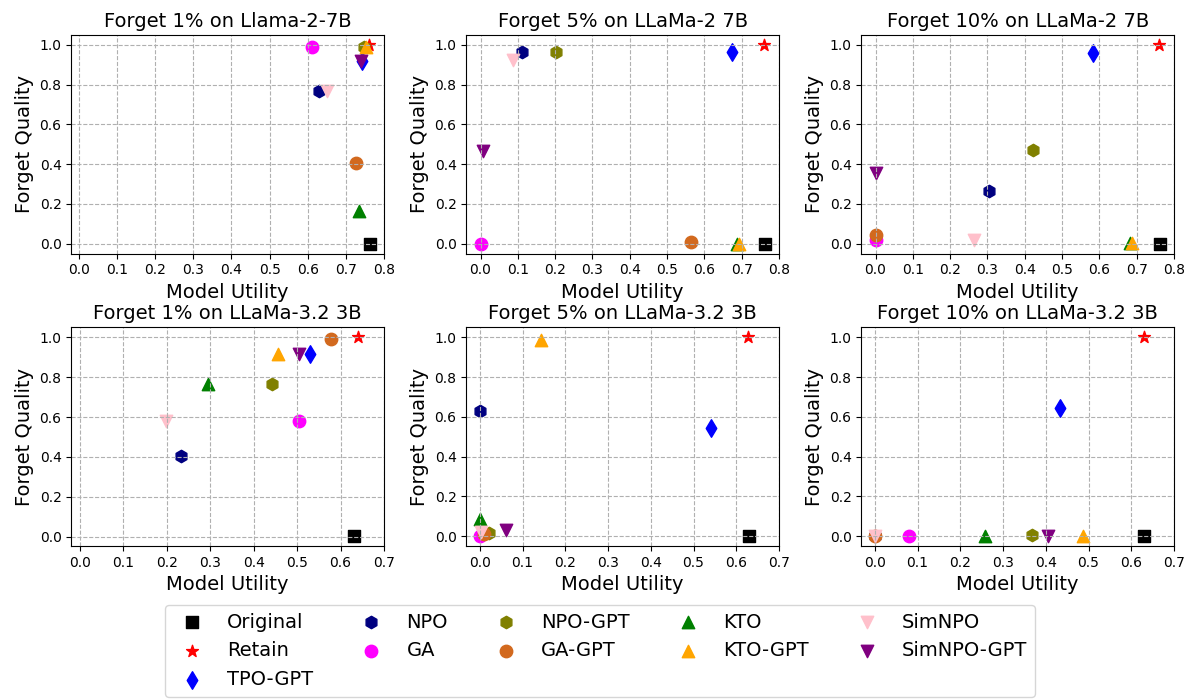}
    \caption{{\textbf{Forget Quality versus Model Utility across varying forget set sizes (1\%, 5\%, and 10\%) after unlearning.} Results are presented for our method \textbf{TPO-GPT} and all baselines, including those incorporating the GPT-based unwanted information identifier. While the identifier improves model utility, all baselines face challenges in maintaining a good balance between forget quality and model utility as the forget set size increases. \textbf{TPO-GPT} demonstrates a notable trade-off. Data points represent the epoch at which each method achieves its peak forget quality.}}
    \label{fig:tofu_result}
\end{figure*}

\begin{table}[t]
    \centering
    \small
    \setlength{\tabcolsep}{1mm} 

    \begin{tabular}{c|ccc|c}
        \toprule
        \multirow{3}{*}{\bf Method} & \multicolumn{3}{c|}{\bf Forget Quality} & {\bf Model Utility} \\ 
        
        & VerbMem & KnowMem & PrivLeak & KnowMem \\ 
        & $\mathcal{D}_f (\downarrow)$ & $\mathcal{D}_f (\downarrow)$ & $ (\rightarrow0)$ & $\mathcal{D}_r (\uparrow)$ \\
        \midrule
         \multicolumn{5}{c}{\bf MUSE News}\\
        \midrule
        Original   & 56.26 & 63.66 & -99.81 & 54.63 \\
        Retain    & 19.83 & 31.73 & 0.00  & 55.25 \\
        \midrule
        GA         & 0.00     & 0.00     & 20.24 & 0.00 \\
        NPO        & 0.00     & 0.00     & 18.57 & 0.00 \\
        SimNPO        & 0.00     & 2.12     & 2.80 & 0.00 \\
         TPO        & 0.00  & 0.00     & \bf 2.60 & 0.00 \\
        \midrule
         Task Vector & 66.74 & 62.53 & -100 & \bf 50.28 \\
        GA$_{\text{GDR}}$      & 4.89     & 21.18     & 109.56 & 5.85 \\
        NPO$_{\text{GDR}}$     & 0.00     & 45.02     & 109.56 & 42.37 \\
        SimNPO$_{\text{GDR}}$        & 35.32      & 53.03     & -97.17 & 45.82 \\
         TPO$_{\text{GDR}}$     & 29.38  & 54.67  & \bf -6.12 &  43.67 \\
        \midrule
         \multicolumn{5}{c}{\bf MUSE Books}\\
        \midrule
        Original   & 99.70 & 45.87 & -57.14 & 69.40 \\
        Retain    & 13.88 & 30.13 & 0.00  & 69.04 \\
        \midrule
        GA         & 0.00     & 0.00     & -23.23 & 0.00 \\
        NPO        & 0.00     & 0.00     & -23.75 & 0.00 \\
        SimNPO     & 0.00     & 0.00     & \bf -10.60 & 1.16 \\
         TPO        & 0.15  & 0.00     & -19.50 & 0.00 \\
        \midrule
        Task Vector & 98.94 & 41.63 & -76.97 & \bf 67.18 \\
        GA$_{\text{GDR}}$      & 0.00     & 0.00     & -24.19 & 3.74 \\
        NPO$_{\text{GDR}}$     & 0.00     & 0.00     & -27.86 & 10.57 \\
        SimNPO$_{\text{GDR}}$        & 0.00     & 1.62     & -25.81 & 52.69\\
         TPO$_{\text{GDR}}$     & 5.20  & 3.79  & \bf -20.66  & 42.07 \\
        \bottomrule
    \end{tabular}%
    \caption{ Forget quality and model utility for various methods on the MUSE dataset using LLaMA-2 7B. Large positive/negative PrivLeak values indicate over/under-unlearning. {\bf Bolded} results represent the best performance.}
    \label{tab:muse_results}
\end{table}

\section{Results and Discussion}
\subsection{Performance on TOFU}\label{subsec:tofu_result}
\textbf{Unwanted information identification enhances unlearning performance.} We present the unlearning performance of baseline methods, i.e., GA, NPO, KTO, and SimNPO, along with those enhanced by the proposed unwanted information identifier using GPT, i.e., GA-GPT, NPO-GPT, KTO-GPT, and SimNPO-GPT.  \textbf{TPO-GPT} is our proposed method in this work. Figure~\ref{fig:tofu_result} clearly shows that methods utilizing the unwanted information identification consistently demonstrate a superior model utility while achieving a comparable level of forget quality in most scenarios. Notably, for smaller forget set sizes (e.g., 1\%), the unwanted information identifier also enhances the forget quality of baseline methods like NPO-GPT, KTO-GPT, and SimNPO-GPT. These results underscore the effectiveness of selectively unlearning unwanted information while preserving general information.

\textbf{TPO-GPT achieves the best forget quality on a larger forget set size.} Figure~\ref{fig:tofu_result} illustrates that all baseline methods experience a significant decline in forget quality as the forget set size increases. Notably, at a forget set size of 10\%, GA-based and KTO-based methods fail completely in unlearning for both LLaMA2 7B-Chat and LLaMA3.2 3B models, evidenced by their near-zero forget quality. Further, while NPO-based and SimNPO-based methods achieve higher forget quality, their performance also noticeably declines when the forget set size reaches 10\%.

In contrast, the developed TPO-GPT consistently demonstrates comparable forget quality on smaller forget set sizes (e.g., 1\% and 5\%) and achieves optimal forget quality on larger forget set sizes (e.g., 10 \%) for both models. Notably, on LLaMA2 7B-Chat, TPO-GPT consistently achieves forget quality exceeding 90\% across different forget set sizes, as evidenced by the first row of Figure~\ref{fig:tofu_result}. 

\textbf{TPO-GPT preserves utility while achieving the best trade-off under larger forget set sizes.} As shown in Figure~\ref{fig:tofu_result}, TPO-GPT consistently maintains high model utility and strong forget quality, even as the forget set size increases. At 1\% and 5\%, it achieves over 85\% utility and near-perfect forget quality on both LLaMA-2 7B and LLaMA-3.2 3B. Notably, under the most challenging condition of forgetting 10\%, TPO-GPT still preserves 70\% utility while maintaining the highest forget quality among all methods. These results highlight TPO-GPT's effectiveness in balancing unlearning performance and model preservation, especially under demanding unlearning scenarios.

\subsection{Performance on MUSE}
\textbf{GDR significantly improves the utility preservation.} As shown in the Table~\ref{tab:muse_results}, nearly all unlearning methods suffer from severe utility degradation on the MUSE benchmark when GDR is not used. This is largely attributed to the large size of the forget set in the MUSE Benchmark, which poses a challenge to preserving general model performance. Incorporating GDR consistently mitigates this issue and improves utility across all methods.

\noindent
\textbf{TPO\textsubscript{GDR} consistently achieves the best PrivLeak performance.} 
PrivLeak serves as the primary metric to measure the performance gap from the retained model defined in Section~\ref{lab:retained_model}. Notably, our TPO\textsubscript{GDR} method consistently achieves PrivLeak values closest to 0 on both News (-6.12) and Books (-20.66), while maintaining comparable KnowMem and VerbMem scores on the forget set relative to other GDR-based baselines. Although Task Vector preserves high model utility on both tasks, it fails completely in unlearning, as its forget quality remains nearly identical to that of the original model. Overall, TPO offers the best trade-off between effective unlearning (lowest PrivLeak) and reasonable utility retention, showing superior performance over all baselines.

\section{Related Work}
{\bf LLM Unlearning.} Motivated by data privacy regulations like the General Data Protection Regulation (GDPR) that gave individual users the ``right to be forgotten''~\cite{rosen2011right}, machine unlearning was initially developed to remove the effect of specific training examples without retraining the model on the entire data~\cite{cao2015towards,bourtoule2021machine}. Its effectiveness has been shown in different domains, including image classification~\cite{sekhari2021remember,fan2025challenging} and federated learning~\cite{wang2022federated,jin2023forgettable}. However, these unlearning methods often become infeasible for LLMs due to the massive parameter sizes in LLMs.

Recent efforts have adapted unlearning to LLMs by fine-tuning with gradient ascent on the forget set and gradient descent or KL divergence on the retain set~\cite{li2024wmdp,yao2023large,chen2023unlearn,jang2022knowledge,wang2023kga,tian2024forget,liu2024towards,ji2024reversing,wang2024llm,zhao2024makes,jia2024soul}.
Yet, existing methods~\cite{yao2023large,liu2022continual} often struggle to balance forgetting and utility preservation, frequently resulting in catastrophic collapse, as observed on benchmarks like TOFU~\cite{maini2024tofu}. To address this limitation, NPO~\cite{zhang2024negative}, inspired by direct preference optimization (DPO)~\cite{rafailov2024direct}, introduces a lower-bounded unlearning objective to mitigate catastrophic collapse. Additionally, Simple Negative Preference Optimization (SimNPO)~\cite{fan2024simplicity} enhances NPO by proposing a reference-free variant, drawing inspiration from Simple Preference Optimization (SimPO)~\cite{meng2024simpo}. However, the performance of these methods deteriorates as the size of the forget set increases~\cite{liu2024large}, underscoring the need for more robust and scalable solutions to achieve effective unlearning while preserving model utility.

\noindent
{\bf Targeted Unlearning.} Recent work, such as RESTOR~\cite{wan2025not,liu2024revisiting,rezaei2024restor}, demonstrates that isolating and precisely targeting the information within the unlearning scope significantly improves the unlearning performance of existing methods (e.g., GA \cite{yao2023large}, WHP~\cite{eldan2023s}), highlighting the crucial role of targeted unlearning. However, a major challenge remains: knowledge dependencies~\cite{liu2024large} make it difficult to cleanly separate the information that should be forgotten from what should be retained. Recent methods tackle this challenge through varied strategies: MemFlex \cite{tian2024forget} leverages gradient information to focus on sensitive parameters accurately. ECO~\cite{liu2024large} proposes an efficient unlearning framework that localizes unlearning to sentences that contain content within the unlearning target by using a sentence-level identifier. However, \cite{thaker2024position} shows that ECO's classifier tends to overfit specific keywords rather than align with the unlearning requester's intent, leading to insufficient unlearning when those keywords are absent or rephrased in the sentence. Additionally, SEUL~\cite{wang2024selective} achieves fine-grained information identification by incorporating a sensitive span annotation framework that uses an LLM (e.g., ChatGPT)  to annotate specific spans containing sensitive information. This approach improves unlearning effectiveness by targeting specific continuous sequence spans. However, it focuses solely on PII unlearning, overlooking broader generalizability to various unlearning tasks. (e.g., copyrighted content unlearning). Despite these advancements, challenges such as over- and under-forgetting remain, highlighting the need for more precise and robust solutions to disentangle information dependencies between forget and retention sets~\cite{thaker2024position,liu2024rethinking,zhao2024makes}. 

\section{Conclusion~\label{sec:conclusion}}
In this work, we propose TIF, a framework that improves LLM unlearning by distinguishing between UW and GW. TIF employs the TPO objective to selectively unlearn UW while preserving GW. Experiments on TOFU and MUSE benchmarks show that TIF enhances unlearning effectiveness for existing unlearning methods and substantially preserves more model utility. Our study focuses on {\it sequence unlearning} by suppressing token generation, relevant to copyright and privacy protection. In contrast, knowledge unlearning (e.g., WMDP~\cite{li2024wmdp}) targets unlearning entire distributions of hazardous knowledge from latent representations, advancing model security in domains such as biosecurity, cybersecurity, and chemical safety.

\section{Ethical Statement}
This work develops methods for targeted unlearning in large language models to enhance privacy and compliance with data protection laws. All experiments use public datasets (TOFU and MUSE) without personal or sensitive data. We highlight that unlearning techniques should be applied responsibly to prevent misuse or selective information removal.

\section{Acknowledgments}
This work was supported in part by the National Science Foundation under Awards 2504264, 2211897, 2211896, and 2401775.

\bibliography{main}

\clearpage
\appendix
\setcounter{secnumdepth}{2} 
\renewcommand{\thesubsection}{\thesection.\arabic{subsection}}

\section{Appendix: Additional Experiment Details\label{sec:detail}}

\noindent
\subsection{Computational Configurations} All experiments are conducted on 2 NVIDIA H100 GPU cards in a single node.

{\bf (1) MUSE.}
We use LLaMA-2 7B fine-tuned on BBC news articles as the original model for News and ICLM-7B fine-tuned on Harry Potter books as the original model for Books. For unlearning, we trained the model for 10 epochs with a learning rate fixed at $1e^{-5}$ and a batch size of 32. 

We utilize the default setting for NPO with the value of parameter $\beta$ fixed at 0.1. For TPO, we introduce a tuning weight for PL, formally:
\begin{small}
    \begin{align*}
\mathbb{E}_{\xi_f\sim D_f}\bigg[ \! \underbrace{-\frac{2}{\beta}\log\sigma\big(\beta  (z_{\text{\bf ref}}(\hat{y}|x_f) - z_{\boldsymbol{\theta}}(\hat{y}|x_f)\big)}_{\text{LPL}}  - \underbrace{\lambda \log P_{\boldsymbol{\theta}}(\bar{y} | x_f)}_{\text{PL}} \!\bigg].
 \nonumber
\end{align*}
\end{small}We conduct a grid search for $\beta$ in a range of [0.1,0.3] and for $\lambda$ in a range of [0,0.01]. The optimal $\beta$ values, which deliver the best unlearning performance when TPO and TPO$_{\text{GDR}}$ achieve the best forget quality across various tasks and models, are presented in Table~\ref{tab:muse_beta}.
\begin{table}[h]
    \centering

    \begin{tabular}{c|cc}
    \toprule
     \midrule

 Model& TPO&TPO$_{\text{GDR}}$\\
     \midrule
         LLaMa-2 7B (News)& 
          0.2&0.2\\
 ICLM-7B (Books)& 0.2&0.2\\
     \midrule
        \bottomrule
    \end{tabular}
    \caption{Optimal $\beta$ values when TPO and TPO$_{\text{GDR}}$ achieve the best forget quality across different models and tasks in the MUSE benchmarks (News and Books).}
    \label{tab:muse_beta}
\end{table}

{\bf (2) TOFU.}
In all experiments, the models are trained using the AdamW optimizer with a weight decay of 0.01. A linear warm-up is applied during the first epoch, with the learning rate fixed at $1e^{-5}$ and the batch size of 32. The original model is fine-tuned on TOFU for 5 epochs. Unlearning is performed on the initial model for 10 epochs using our TPO method and all baseline methods.

For unlearning with NPO, we use the default setting, fixing the parameter $\beta$ at 0.1. For TPO, the parameter $\beta$ is tuned by searching within the range [0.1, 0.5] to obtain the best-performing model. We report the value of $\beta$ that yielded the best unlearning performance when TPO achieves the best forget quality across different tasks and models in 

\subsection{Computational Efficiency of the Unwanted Information Identification} 
Distinguishing between UW from GW using either generative or discriminative approaches remains computationally efficient and time-effective. For smaller data sets like TOFU, using a generative LM like ChatGPT-4o to process it takes several minutes. For larger datasets like MUSE, using a discriminative model like DistilBERT on our H100 server completes the task in just a few hours. These customized approaches highlight our method's flexibility, computational efficiency, and scalability across diverse dataset sizes.

\subsection{Evaluation Metrics}
We summarize the LLM models and the evaluation metrics used across various unlearning benchmarks in Table~\ref{tab:summary}.

\subsection{Training Cost Analysis}
\label{sec:training_cost}
We evaluate the computational overhead of our proposed TPO-GPT compared to the baseline NPO-GPT. TPO-GPT requires only about $1.7\times$ more training time and incurs roughly $6\%$ higher GPU memory consumption. These results indicate that TPO introduces moderate additional cost while providing substantial unlearning benefits.

\section{Additional study on individual components of TPO}
Our TPO loss comprises two components: the LDL and the PL. To further disentangle their individual contributions, we conduct an ablation study on the Forget-05 subset of TOFU. The role of PL in maintaining model utility has been discussed in Section~\ref{sec:TPO}, and its effectiveness is further demonstrated by the results in Table~\ref{tab:LDL}. Both LDL and NPO exhibit reduced utility degradation when combined with the PL loss. Moreover, LDL plays a central role in TPO by effectively balancing forget quality and model utility. Compared to NPO variants, LDL alone preserves substantially more utility, and when integrated with PL and our GPT-based unwanted information identification process, it achieves the best trade-off between FQ and MU. These findings underscore the complementary roles of LDL and PL in enabling targeted unlearning.

\begin{table}[t]
    \centering
    \begin{tabular}{c|ccc}
        \toprule
        \midrule
         \multirow{2}{*}{Model}&   \multicolumn{3}{c}{TPO}\\
 &Forget 01& Forget 05& Forget 10\\
 \midrule
         LLaMa-2 7B & 
          0.32& 0.32& 0.23\\
 LLaMa-3.2 3B& 0.3& 0.27& 0.19\\
        \midrule
        \bottomrule
    \end{tabular}
    \caption{Optimal $\beta$ values when TPO achieves the best forget quality across different models and tasks in the TOFU benchmarks.}
    \label{tab:tofu_beta}
\end{table}

\begin{table*}[t]
    \centering
    \begin{tabular}{c|cccccc}
    \toprule
    \midrule
         Forget 05&   NPO  &NPO-GPT& NPO+PL-GPT
& LDL-only & LDL-GPT&TPO (LDL+PL)-GPT 
\\
         \midrule
         FQ& 0.96 &0.96 & 0.96
& 0.55 & 0.92&\bf 0.96 
\\
         MU& 0.11 &0.21& 0.42& 0.51 & 0.63&\bf 0.67 \\
         \midrule
    \bottomrule
 
     \end{tabular}
    \caption{Comparison of unlearning performance on the Forget-05 subset of TOFU between LDL (a core component of TPO) and NPO methods. ``F'' denotes forget quality, and ``MU'' denotes model utility.}
    \label{tab:LDL}
\end{table*}

\begin{table*}[t]
    \centering
        \begin{tabular}{c|c|cc|cc}
            \toprule
            \midrule
                 Benchmark &  Used LLM & \multicolumn{2}{c|}{Forget quality} & \multicolumn{2}{c}{Model Utility} \\
                 \cmidrule{1-6}
                 \multirow{3}{*}{MUSE} &  \multirow{2}{*}{ICLM-7B} & KnowMem on $\mathcal{D}_f$ & $\downarrow$ & \multirow{3}{*}{KnowMem on $\mathcal{D}_r \quad \uparrow$} & \\
                 & & VerbMem on $\mathcal{D}_f$ & $\downarrow$ & & \\
                 & LLaMa-2 7B & PrivLeak & $(\rightarrow 0)$ & & \\
                \cmidrule{1-6}
                 \multirow{3}{*}{TOFU} & \multirow{2}{*}{LLaMa-2 7B} & \multirow{3}{*}{Truth Ratio on $\mathcal{D}_f$} & \multirow{3}{*}{$\uparrow$} & \multirow{3}{*}{$\text{Mean} \Bigg(
                \begin{cases} 
                    \text{Probability, Rouge-L, Truth Ratio} \\
                    \mathcal{D}_r, \mathcal{D}_{\text{real\_authors}}, \mathcal{D}_{\text{word\_facts}}
                \end{cases}
              \!\!\!\!\!\!\!   \Bigg)$}
                & \multirow{3}{*}{$\uparrow$} \\
                 & & & & & \\
                 &  LLaMa-3.2 3B & & & & \\     
                 \midrule
                 \bottomrule
         \end{tabular}
    \caption{ Summary of unlearning evaluation metrics and used models across different benchmarks.}
    \label{tab:summary}
\end{table*}

\section{Unlearning performance combining the GDR with the proposed TIF framework.}
 Table~\ref{tab:tofu_05_results} summarizes the forget quality and model utility of our TPO compared to several baseline approaches (i.e., GA, KTO, NPO, SimNPO), evaluated on the TOFU Forget 05 task. The experiments are conducted under two conditions: with and without incorporating Gradient Descent on Retain (GDR) loss, and with and without our GPT-based unwanted information identifier (GPT). Here, "Vanilla" denotes methods evaluated without incorporating the GPT-based identifier.

The integration of our GPT-based unwanted information identifier consistently enhances both forget quality and model utility across most baseline methods, demonstrating the effectiveness of our proposed framework in accurately distinguishing unwanted information from general knowledge. Notably, our TPO method achieves the highest forget quality on the LLaMa-2 7B model and preserves substantially more model utility compared to all other baseline methods across all experimental conditions. These results highlight TPO's effectiveness in unlearning and underscore the importance of accurate unwanted information identification in maintaining model utility.

\begin{table*}[t]
\centering
\begin{tabular}{c|cccc|cccc}
\toprule
\midrule
\bf Method & \multicolumn{4}{c|}{\bf LLaMa-3.2 3B}& \multicolumn{4}{c}{\bf LLaMa-2 7B}\\
 & \multicolumn{2}{c}{{\bf Forget Quality}}& \multicolumn{2}{c|}{{\bf Model Utility}}& \multicolumn{2}{c}{{\bf Forget Quality}}& \multicolumn{2}{c}{{\bf Model Utility}}\\
\midrule
Original   & \multicolumn{2}{c}{0}& \multicolumn{2}{c|}{0.63}& \multicolumn{2}{c}{0}& \multicolumn{2}{c}{0.76}\\
Retain    &  \multicolumn{2}{c}{1}& \multicolumn{2}{c|}{0.69}& \multicolumn{2}{c}{1}& \multicolumn{2}{c}{0.76}\\
\midrule
 & Vanilla& GPT& Vanilla& GPT& Vanilla& GPT& Vanilla&GPT\\
\midrule
GA& 0.00 & 0.01 \mytaggreen{$\uparrow$} & 0.00 & 0.01 \mytaggreen{$\uparrow$} & 0.00 & 0.01\mytaggreen{$\uparrow$} & 0.00 & 0.57 \mytaggreen{$\uparrow$}\\
 KTO& 0.09 & \bf 0.98 \mytaggreen{$\uparrow$} & 0.00 & 0.14 \mytaggreen{$\uparrow$} & 0.00 & 0.00 \mytagblue{$\sim$} & 0.68 & \bf0.69 \mytaggreen{$\uparrow$}\\
 NPO& 0.63 & 0.02 \mytagred{$\downarrow$} & 0.00 & 0.02 \mytaggreen{$\uparrow$} &0.96  & \bf 0.96  \mytagblue{$\sim$}& 0.11 & 0.21 \mytaggreen{$\uparrow$}\\
SimNPO     & 0.02   & 0.03 \mytaggreen{$\uparrow$}& 0.00  & 0.06 \mytaggreen{$\uparrow$} & 0.92  & 0.63 \mytagred{$\downarrow$} & 0.08  & 0.36 \mytaggreen{$\uparrow$}\\
 TPO     & - & 0.54\quad${}$ & - & \bf0.54\quad${}$   & - & \bf 0.96\quad${}$ & - & 0.67\quad${}$  \\
\midrule

 GA$_{\text{GDR}}$& 0.00 & 0.00 \mytagblue{$\sim$} & 0.56 & 0.56 \mytagblue{$\sim$} & 0.01 & 0.01 \mytagblue{$\sim$} & 0.43& 0.61 \mytaggreen{$\uparrow$}\\
 KTO$_{\text{GDR}}$& 0.07 & 0.01 \mytagred{$\downarrow$} & 0.0 & 0.02 \mytaggreen{$\uparrow$}& 0.00 & 0.01 \mytaggreen{$\uparrow$}& 0.73 & 0.08 \mytagred{$\downarrow$}\\
 NPO$_{\text{GDR}}$& 0.22 & \bf 0.71 \mytaggreen{$\uparrow$} & 0.60 & 0.44 \mytagred{$\downarrow$} & 0.22 & 0.79 \mytaggreen{$\uparrow$} & 0.56 & 0.56 \mytagblue{$\sim$}\\
 SimNPO$_{\text{GDR}}$     & 0.00   & 0.01\mytaggreen{$\uparrow$} & 0.61  & \bf 0.64 \mytaggreen{$\uparrow$}& 0.00 & 0.07 \mytaggreen{$\uparrow$} & 0.71  & \bf0.72 \mytaggreen{$\uparrow$}\\
  TPO$_{\text{GDR}}$     &  - & 0.55\quad${}$ & - & 0.61\quad${}$ & - & \bf 0.80\quad${}$ & - & 0.70\quad${}$ \bf   \\
 \midrule
\end{tabular}%
\caption{ Forget quality and model utility for our TPO method and various baselines evaluated on the TOFU Forget 05 task. Results are presented with and without incorporating Gradient Descent on Retain (GDR) loss and our GPT-based unwanted information identifier (GPT). Improvements achieved by incorporating the GPT-based identifier compared to methods without it (denoted as Vanilla) are marked as \mytagred{$\uparrow$}, similar performances as \mytagblue{$\sim$}, and declines as \mytaggreen{$\downarrow$}. Best performances are \textbf{boldfaced}.}
\label{tab:tofu_05_results}
\end{table*}

\section{Generative LM Approach vs Discriminative LM Approach.\label{subsec:ga_vs_da}} 
\begin{figure}[t]
    \centering
    \includegraphics[width=1\linewidth]{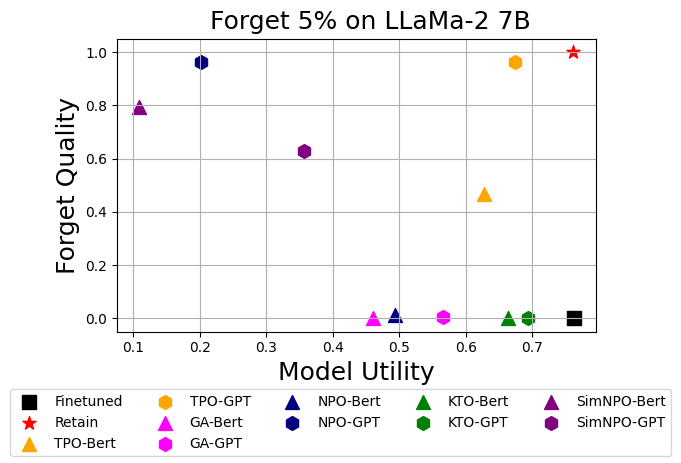}
    \caption{ {\bf Forget Quality versus Model Utility on the Forget05 Task.} The figure compares our method \textbf{TPO} with various baselines, integrating generative LM-based (GPT) and discriminative LM-based (DistilBERT) information identifiers. Hexagonal markers denote results from the unlearning methods using the GPT, while triangle markers correspond to the methods using DistilBERT.}
    \label{fig:bert-vs-GPT}
    \vspace{-0.4cm}
\end{figure}

We compare the unlearning effectiveness of unlearning methods that separately incorporate generative LM-based and discriminative LM-based identifiers. As shown in Figure~\ref{fig:bert-vs-GPT}, generative LM-based (GPT) methods (hexagonal markers) consistently achieve higher forget quality compared to discriminative LM-based (Bert) methods (triangular markers). Additionally, GPT-based methods such as GA-GPT, KTO-GPT, SimNPO-GPT, and TPO-GPT preserve more utility. These results confirm the generative LM approach's superior effectiveness in unlearning compared to the discriminative approach. Among these, our TPO-based method achieves the best trade-off between forget quality and model utility, regardless of the identifier type.

\section{Baselines}\label{sec:baselines}
In this section, we outline and analyze the baseline methods used for comparison in our experiments. These methods represent established approaches in the field of machine unlearning and serve as benchmarks for evaluating the effectiveness of our proposed method.
\subsection{Gradient ascent} 
Gradient ascent (GA) is a fundamental technique in many existing machine unlearning works~\cite{yao2023large,jang2022knowledge} that prevents generating undesirable texts using only negative samples. The GA loss is shown as follows:
\begin{small}
\begin{align*}
    \ell_{\text{GA}}(\boldsymbol{\theta}) \coloneqq \mathbb{E}_{\xi_f \sim D_{f}}\log \big(P_{\boldsymbol{\theta}}(y_f | x_f) \big),
\end{align*}
\end{small}where $P_{\boldsymbol{\theta}}(y|x)$ is the predicted probability of generating a sequence of tokens $y$ by an LLM $\mathcal{M}_{\boldsymbol{\theta}}$ conditioned on the prompt $x$.

In gradient ascent, the objective is to increase the prediction loss on $\mathcal{D}_f$ by inverting the direction of the cross-entropy objective. This technique often proves effective for smaller datasets and is typically limited to a few training epochs to prevent the model from collapsing into trivial solutions that degrade its overall utility~\cite{tian2024forget}. However, if applied over a prolonged unlearning process, 
It can lead to catastrophic collapse, causing the model’s utility to degrade drastically, rendering it practically unusable~\cite{zhang2024negative}.

\subsection{Negative preference optimization (NPO)}
To tackle the catastrophic collapse, NPO~\cite{zhang2024negative} redefines the preference optimization process to better align with unlearning objectives, by focusing solely on penalizing undesired outputs without requiring corresponding positive feedback. This approach ensures that the model suppresses the likelihood of generating sensitive or unwanted content from the forget set 
$\mathcal{D}_f$, effectively mitigating the risk of collapse while preserving overall utility in safe response generation. The NPO loss is defined as:
\begin{small}\begin{align*}
\ell_{\text{NPO}}(\boldsymbol{\theta}) \! \coloneqq \! -\frac{2}{\beta} \mathbb{E}_{\xi_f \sim D_{f}} \! \biggl[ \log\sigma \bigg(\!\!\!-\beta \log \bigg(\!\frac{P_{\boldsymbol{\theta}}(y_f \mid x_f)}{P_{\boldsymbol{\theta}_{\boldsymbol{\theta}_\text{o}}}(y_f \mid x_f)}\!\bigg) \!\bigg)\! \biggr],
\end{align*}\end{small}where $\sigma(t) = 1/(1+e^{-t})$ is a sigmoid function, $\beta > 0$ is the inverse temperature. The term $P_{\text{ref}}(y_f|x_f)$ denotes the probability assigned to token $y_f$ given an input $x_f$ by the original model $\mathcal{M}_{\boldsymbol{\theta}_\text{o}}$, whose parameters remain frozen during unlearning.

The NPO loss addresses the issue of catastrophic collapse by introducing a lower bound that holds for any finite $\beta > 0$, thereby ensuring a more stable unlearning process.

\subsection{Kahneman-Tversky optimization (KTO)}
We consider KTO~\cite{ethayarajh2024kto} as another baseline method. KTO is an alignment technique that relies solely on non-paired preference data. Following \cite{zhang2024negative}, we employ the same modified variant of the original KTO. The KTO loss is given as follows:
\begin{small}\begin{align*}
   & \ell_{\text{KTO}}(\boldsymbol{\theta}) \!  \coloneqq \! - \frac{2}{\beta} \mathbb{E}_{\xi_f \sim D_f} \! \biggl[\log \sigma \bigg(\!\! KL_{\boldsymbol{\theta}_\text{o}} - \beta\log \bigg( \frac{P_{\boldsymbol{\theta}}(y_f | x_f)}{P_{\boldsymbol{\theta}_\text{o}}(y_f|x_f)} \! \bigg)  \! \bigg) \! \biggl], \\
   & KL_{{\boldsymbol{\theta}_\text{o}}}(\boldsymbol{\theta})  \coloneqq \mathbb{E}_{\xi_f \sim D_f} \big[ \beta \cdot KL(P_{\boldsymbol{\theta}}(y_{\textnormal{safe}}|x_f)||P_{\boldsymbol{\theta}_\text{o}}(y_{\textnormal{safe}}|x_f)) \big],
\end{align*}\end{small}where $y_{\textnormal{safe}} \coloneqq $ ``I don't know'', $\beta>0$ is the inverse-temperature, and $\sigma$ is the sigmoid function. Compared to NPO, KTO incorporates an additional ``I don't know'' response for unrelated outputs, enhancing unlearning by aligning it closely with human preferences for specific tasks.

\subsection{Task Vectors}
Task Vectors~\cite{ilharco2022editing} provides an efficient mechanism for modifying neural network behavior through simple arithmetic on model weights, making them particularly effective for unlearning tasks. The process begins by fine-tuning the original model $\mathcal{M}_{{\boldsymbol{\theta}}_0}$ on forget set $\mathcal{D}_f$ until the model overfits, producing a reinforced model $\mathcal{M}_{\text {reinforce}}$. A task vector is then computed to capture the difference in weight updates between the original model and the reinforced model, formally defined as: $\mathcal{M}_{{\boldsymbol{\theta}}_0}$ and $\mathcal{M}_{\text {reinforce}}$, where formally: $\Delta \mathcal{M} = \mathcal{M}_{\text {reinforce}} - \mathcal{M}_{{\boldsymbol{\theta}}_0}$. To achieve unlearning, this $\Delta \mathcal{M}$ is subtracted from the original model's weights. Formally:
\begin{equation*}
\mathcal{M}_{\boldsymbol{\theta}} = \mathcal{M}_{{\boldsymbol{\theta}}_0} - \Delta \mathcal{M}.
\end{equation*}This approach intuitively drives the model parameters away from the trajectory induced by the $\mathcal{D}_f$, enabling the effective removal of learned information while preserving the general utility of the original model.

\begin{table*}[t]
    \centering
            \begin{tabular}{l}
               \toprule
                \hline
                 \multirow{2}{*}{\large \textbf{Instruction for Chat-GPT: Identifying Unwanted Words in TOFU}}\\  \\
                 \hline
                \textbf{1. Identify Important Words for All Question and Answer Pairs:} \\
                   \textbullet\quad For each question and answer pair provided, identify the important words. \\
                  \textbullet\quad If the question \textbf{explicitly asks for the author's name}, include the author's name as an important word in the answer. \\
                  \textbullet\quad If the question \textbf{does not ask for the author's name}, exclude the author's name and focus on the other key words \\
                  \quad\ \  in the answer.\\
                 \textbf{2. Key Words to Include: } \\
                  \textbullet\quad Important words should \textbf{directly answer the question and be sufficient to provide a complete and exact answer. }\\
                  \textbullet\quad The selected words should be: \\
                  \quad \textasteriskcentered \quad Proper nouns (excluding author names if not specifically asked). \\
                  \quad \textasteriskcentered \quad Technical terms, specific concepts, or notable features that address the main details of the question. \\
                  \quad \textasteriskcentered \quad \textbf{Specific roles, occupations, places, or other information} that directly contribute to the answer.\\
                 \textbf{3. Key Words to Exclude: }\\
                  \textbullet\quad Do \textbf{not} include words that are \textbf{contextual} but do not directly contribute to answering the question (e.g.,   \\ 
                  \quad \ \  ``father" or ``mother" if the question asks for their specific occupations).\\
                 \textbf{4. Output Format:  }\\
                  \textbullet\quad Provide the results directly in the response. \\
                  \textbullet\quad For each question-answer pair, include a target\_words attribute. \\
                  \textbullet\quad The target\_words attribute should be a \textbf{list of important words} that \textbf{precisely answer the question.} \\
                  \textbf{5. Example Output Structure: }\\
                  \quad json\\
                  \quad Copy code\\
                  \quad [\\
                 \quad \quad \{\\
                \quad \quad \quad ``question": ``What are the contributions of Albert Einstein?", \\
                 \quad \quad \quad ``answer": ``Albert Einstein made significant contributions to the theory of relativity and quantum  mechanics.", \\
                 \quad \quad \quad ``target\_words": [ ``theory of relativity", ``quantum mechanics" ]\\
                  \quad \quad \}\\
                 
                  \quad  ]\\
                  \quad In this example:\\
                  \quad \quad \textbullet\quad The focus is on \textbf{key details that exactly answer the question.} \\
                  \quad \quad \textbullet\quad Words like ``\textbf{theory of relativity}" and ``\textbf{quantum mechanics}" directly represent Einstein's contributions, and\\
                  \quad \quad \quad\ \   therefore, they are included as target\_words.\\
    \hline
    \bottomrule
    \end{tabular}
    \caption{\textbf{Comprehensive Instructions for Identifying UW (TOFU) using Chat-GPT}: A systematic approach to extracting sensitive or unwanted words from question-answer pairs, focusing on precise and contextually relevant details while excluding extraneous information. Includes clear guidelines, examples, and a structured JSON output format for efficient processing.}
\label{tab:instruction_gpt}
\end{table*}

\section{Additional Experiment on the Generative Language Model Approach \label{sec:instruction_gpt}} We present the task instruction used for identifying unwanted words in the TOFU dataset in Table~\ref{tab:instruction_gpt}. To evaluate the robustness of unwanted word identification using a generative language model approach with ChatGPT-4o, we conducted three experiments with ChatGPT-4o on the TOFU Forget01 set. In all three experiments, the same instructions are used (Table~\ref{tab:instruction_gpt}).

We evaluate the consistency of unwanted word identification by computing the Jaccard index across different runs of our generative approach using ChatGPT-4o. Specifically, we measure pairwise similarity between the unwanted word sets extracted in three independent experiments on the TOFU Forget01 set. The Jaccard index values for these pairwise comparisons are 0.887, 0.909, and 0.869, demonstrating that our instruction design ensures stable and consistent identification of unwanted words within the TOFU dataset.

\section{Limitations\label{sec:limitations}}
While our TIF framework advances traditional sequence-level unlearning by operating at a targeted token-level granularity, its effectiveness relies on the accuracy of the unwanted information identifier. This design may not hold in settings where the unlearning target is conceptually diffuse or implicitly represented in the model’s knowledge, as in benchmarks like WMDP, which emphasize knowledge-level 
unlearning. In such cases, knowledge is often embedded in the distribution of words rather than localized to specific tokens, making it difficult to identify and unlearn without broader context understanding. Our proposed framework currently focuses on sequence unlearning, where it is easier to identify the specific parts or words associated with the unlearning requester's intent. Future work could explore extending TIF with techniques for knowledge-based identification to address more diffuse, knowledge-level unlearning tasks.

\end{document}